%% file: main.tex
\documentclass{article}

\newcommand{\bnote}[1]{\color{black}#1\color{black}}

\usepackage{microtype}
\usepackage{graphicx}
\usepackage{subfigure}
\usepackage{booktabs} % for professional tables

\usepackage[hyperfootnotes=false]{hyperref}

\newcommand{\rao}[1]{\noindent\textcolor{red}{#1}}

\usepackage{multirow}
\usepackage{array}
\newcolumntype{P}[1]{>{\centering\arraybackslash}p{#1}}
\usepackage[accepted]{icml2024}

\usepackage{amsmath}
\usepackage{amssymb}
\usepackage{mathtools}
\usepackage{amsthm}
\input{commands.tex}

\usepackage[capitalize,noabbrev]{cleveref}

\theoremstyle{plain}

\theoremstyle{definition}

\theoremstyle{remark}

\usepackage[textsize=tiny]{todonotes}

\usepackage{listings}
\begin{document}

\icmltitlerunning{LLM-Modulo Framework for Robust Planning}

%\begin{document}
% "Position Paper:" has to be a part of the title according to the guidelines
\twocolumn[
\icmltitle{Position: LLMs Can't Plan,\\
But Can Help Planning in LLM-Modulo Frameworks}

\begin{icmlauthorlist}
% % LSaldyt: I added the following based on how I parsed Rao's slack messages regarding authorship
\icmlauthor{Subbarao Kambhampati}{asu}
\icmlauthor{Karthik Valmeekam}{asu}
\icmlauthor{Lin Guan}{asu}
\icmlauthor{Mudit Verma}{asu}
\icmlauthor{Kaya Stechly}{asu}
\icmlauthor{Siddhant Bhambri}{asu}
\icmlauthor{Lucas Saldyt}{asu}
\icmlauthor{Anil Murthy}{asu}
%\icmlauthor{Firstname8 Lastname8}{sch}
%\icmlauthor{Firstname8 Lastname8}{yyy,comp}
\end{icmlauthorlist}

\icmlaffiliation{asu}{School of Computing and AI, Arizona State University, Tempe, AZ, USA}
% %\icmlaffiliation{comp}{Company Name, Location, Country}
% %\icmlaffiliation{sch}{School of ZZZ, Institute of WWW, Location, Country}

\icmlcorrespondingauthor{Subbarao Kambhampati}{rao@asu.edu}
% %\icmlcorrespondingauthor{Firstname2 Lastname2}{first2.last2@www.uk}

% % You may provide any keywords that you
% % find helpful for describing your paper; these are used to populate
% % the "keywords" metadata in the PDF but will not be shown in the document
\icmlkeywords{LLMs, Planning, Reasoning}

\vskip 0.3in
]



\printAffiliationsAndNotice{}  % leave blank if no need to mention equal contribution
%\maketitle

\begin{abstract}
%There is considerable confusion about the role of Large Language Models (LLMs) in planning and reasoning tasks. On one side are over-optimistic claims that LLMs can indeed do these tasks with just the right prompting or self-verification strategies. On the other side are perhaps over-pessimistic claims that all that LLMs are good for in planning/reasoning tasks are as mere translators of the problem specification from one syntactic format to another, and ship the problem off to external symbolic solvers. In this position paper, we take the view that both these extremes are misguided. 
We argue that auto-regressive LLMs cannot, by themselves, do planning or self-verification (which is after all a form of reasoning), and shed some light on the reasons for misunderstandings in the literature. We also argue that LLMs should be viewed as universal approximate knowledge sources that have much more meaningful roles to play in planning/reasoning tasks beyond simple front-end/back-end format translators. We present a vision of {\bf LLM-Modulo Frameworks} that combines the strengths of LLMs with external model-based verifiers in a tighter bi-directional interaction regime. We will show how the models driving the external verifiers themselves can be acquired with the help of LLMs. We will also argue that rather than simply pipelining LLMs and symbolic components, this LLM-Modulo Framework provides a  better \textit{neuro-symbolic} approach that offers tighter integration between LLMs and symbolic components, extending the scope of model-based planning/reasoning regimes towards more flexible knowledge, problem and preference specifications. 

\end{abstract}

\setcounter{footnote}{0}

\section{Introduction}
\label{sec:intro}

\begin{figure}[h]
    \centering
    \includegraphics[width=\linewidth]{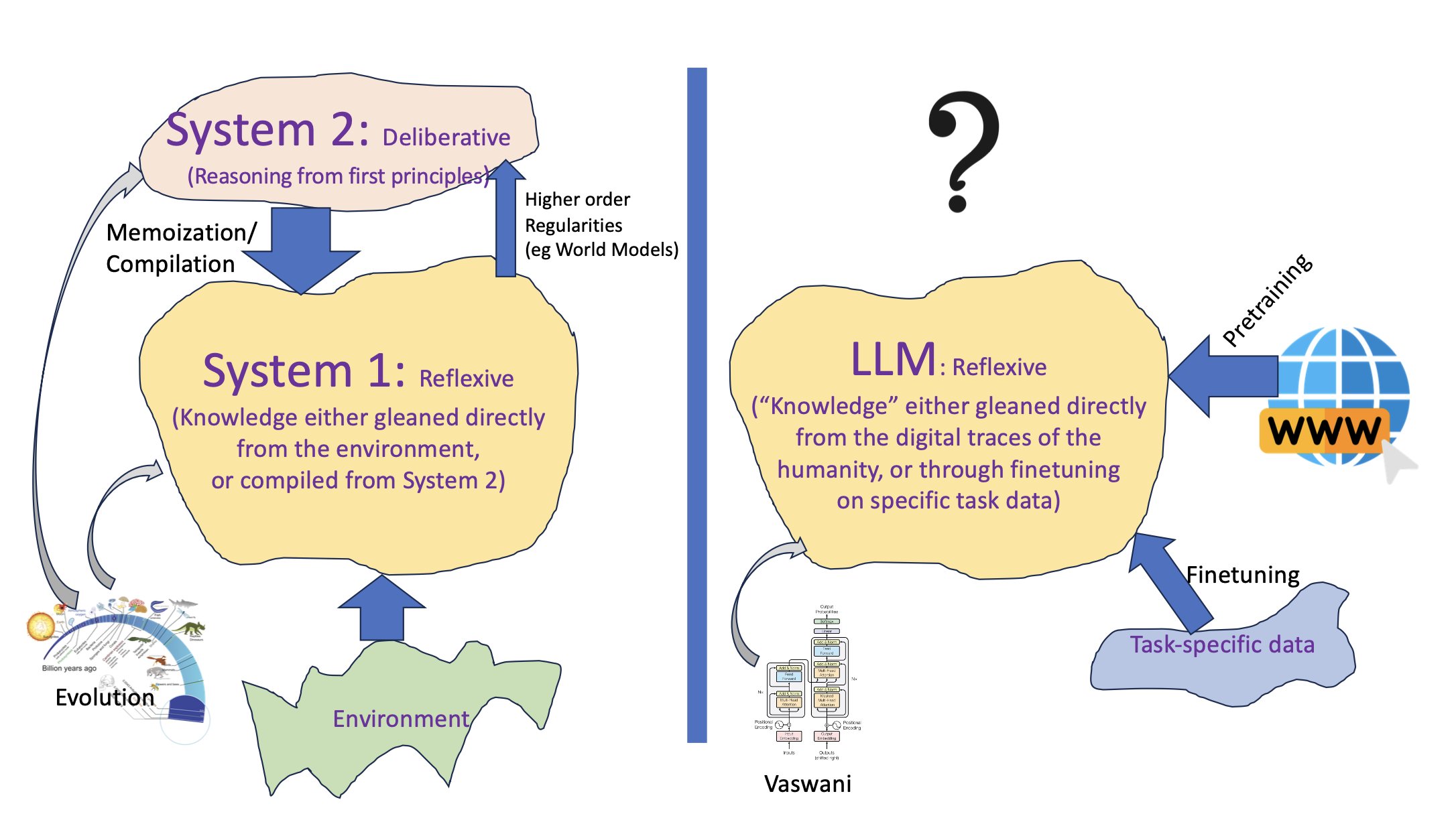}
    \caption{An informal account of viewing an LLM as a giant external non-veridical memory that acts as a pseudo System 1}
    \label{fig:sys12}
\end{figure}

Large Language Models (LLMs), essentially n-gram models on steroids which have been pre-trained on web-scale language corpora
%\rao{Kaya: "corpora"? or "a" before "web-scale"}
(or, effectively, our collective consciousness), have caught the imagination of the AI research community with linguistic capabilities that no one expected text completion systems to possess. Their seeming versatility has led many researchers to wonder whether they can also do well on planning and reasoning tasks typically associated with System 2 competency. On the face of it, this doesn't seem to ring true, as both by training and operation, LLMs are best seen as a giant pseudo System 1 \cite{thinking-fast-slow} (see Figure~\ref{fig:sys12}). Even from a pure engineering perspective, a system that takes constant time to produce the next token cannot possibly be doing principled reasoning on its own.\footnote{Think of asking an LLM an yes/no question--{\em is this theorem logically entailed by this first-order logic knowledge-base}. This is well-known to be a semi-decidable problem. Ask yourself if the LLM will take longer in answering the question.  
(If you are thinking Chain-of-thought prompts or training with step-by-step data, consider that you are essentially changing the nature of the original prompt/training).
%or even the over the top deliberately introducing "Pause" commands doesn't change anything.
}
Not surprisingly, initial excitement based on anecdotal performance of
LLMs on reasoning tasks \cite{bubeck2023sparks} has been dissipated to some
extent by the recent spate of studies, including our own, questioning the robustness of such behaviors--be they planning \cite{valmeekam2023on,rao-cacm}, simple arithmetic and logic \cite{dziri2023faith}, theory of mind \cite{ullman2023large,verma2024theory}, or general mathematical and abstract benchmarks \cite{mccoy2023embers,gendron2023large}.
Despite this, a steady stream of claims continue to be made in the
literature about the planning and reasoning capabilities of LLMs.
In light of questions about their planning capabilities, the head-long rush into agentic LLMs should be particularly concerning. After all, acting without
the ability to plan is surely a recipe for unpleasant consequences!

In an ironic juxtaposition to this unwarranted optimism about the planning and reasoning abilities of LLMs, there is also unwarranted pessimism about the roles LLMs can play in planning/reasoning tasks. Several efforts (e.g. \cite{liu2023llm+,pan2023logic,xie2023translating}) advocate using LLMs only as glorified translators--converting reasoning problems embedded in textual format to symbolic representations, and pawning them off to external classical symbolic solvers (with all their attendant expressivity and search complexity challenges \cite{twotheses-kr}).\footnote{In some circles, this unidirectional pipeline has been given the undeserved badge of \textit{neuro-symbolic architecture}.

%, although one wonders why an LLM extracting and thro
}
%\rao{Anil: Cited Harold Soh's Paper, Ignore below comment}
%\rao{Anil: Should the NUS Harold Soh's Paper be cited above? "Translating Natural Language to Planning Goals with LLMs" https://arxiv.org/abs/2302.05128 - They use LLMs to translate NL goals to PDDL as a central result} 

In truth, LLMs can be a whole lot more than machine translators. They are a kind of approximate knowledge source (albeit sans guarantees) trained on our collective consciousness. While it is unlikely that they will have System 2 competencies by themselves, they can nevertheless be valuable resources in solving System 2 tasks. To put it another way, the problem with Alchemy of yore was not that Chemistry is useless, but that people wanted to delude themselves that Chemistry--a pretty amazing discipline on its own merits--can be Nuclear Physics if you prompt it just so. The confusions regarding LLM abilities, or should we say, LLM alchemy, doesn't seem to be much different--oscillating between ignoring their strengths, and ascribing abilities they don't have.
%\footnote{}

%LLMs can do so much cross-fertilization 

%Mention knowledge sources

%On the face of it, there is no reason to believe that LLMs, which at best 

%\rao{put the t-shirt thing?}

% The delusion comes chiefly from our incessant need to confuse them for
% human intelligence,  merrily applying anthropomorphic concepts such as
% thinking , thoughts, reasoning and self-critiquing to LLMs.
% This anthropomorphization is quite futile, as as we shall see in Section
% 2, even counterproductive and misleading.
% %
% While no one quite foresaw how impressive the approximate omniscience
% of these n-gram models on steroids would be, that doesn't quite justify
% assuming that they do everything humans do.
% %
% If we can manage to tone down the "LLMs are Zero-shot <XXX>" studies
% rife with confirmation biases that conflate approximate retrieval for
% other capabilities associated with human intelligence
% we can focus on the right way of leveraging the strengths of LLMs. This
% can certainly be done in LLM-modulo architectures, with either humans or
% other specialized sound reasoners in the loop).

%\rao{discuss system 1/2}

The goal of this position paper is to introduce some clarity into this confusing state of affairs oscillating between over-optimism and over-pessimism. 
%\fromrao{some of this can go to conclusion too}
Simply put, we take the stance that 
LLMs are amazing giant external non-veridical memories that can serve as
powerful cognitive orthotics for human or machine agents, if rightly
used. The underlying n-gram nature makes them effortlessly intermix what
would be considered disparate fields of study (not surprisingly, LLMs
are seen to be very good at making/finding analogies!). The challenge is
to leverage them without wrongly ascribing to them capabilities they
don't possess. The {\bf LLM-Modulo framework} proposed in this position paper tackles this challenge.

%we argue this is possible. 
%\rao{Lucas: "argue this is possible" in (my opinion, caveat emptor) is weaker than something like "demonstrate this in our proposed neuro-symbolic LLM Modulo architecture"}

For the sake of concreteness, we consider planning tasks, especially as studied in the automated planning community \cite{ghallab2004automated}. 
%Notice that in theory it is possible for LLMs to be very effective as idea generators for external sound planners or humans in the loop in computer-supported cooperative work scenarios, while themselves being very bad at generating plans that are guaranteed to be correct. This is especially likely because the chief power of LLMs comes from their pattern-finding abilities than on first-principles simulations over world models. 
The central position of the paper is that \textit{LLMs cannot plan themselves but can play a variety of constructive roles in solving planning tasks--especially as approximate knowledge sources and candidate plan generators in so-called LLM-Modulo Frameworks, where they are used in conjunction with external sound model-based verifiers}.  
%\rao{Generate-Test with a sound tester will give sound results.}

We support this position by first reviewing literature, including our
own works, that establishes that LLMs cannot be used as planners or plan verifiers themselves (Section~\ref{sec:limitations}). We also discuss why there are claims about planning/verification abilities in the first place, in the process hopefully clarifying some prevalent misunderstandings.

Second, we will propose a framework that allows us to leverage LLMs effectively in planning tasks, by combining them with external critics, verifiers and humans. We call this an {\bf LLM-Modulo Framework} (a name loosely inspired by SAT Modulo Theories \cite{sat-modulo}); see Figure~\ref{fig:llm-modulo}. LLMs play a spectrum of roles in this architecture, from guessing candidate plans, to translating those plans into syntactic forms that are more accessible to external critics, to helping end users flesh out incomplete specifications, to helping expert users acquire domain models (that in turn drive model-based critics). All this leveraging of LLMs is done without ascribing to them any planning or verification abilities. The LLM ideas are vetted by external critics, thus ensuring that the plans generated in this architecture can have formal correctness guarantees where possible.

\section{Planning-centered Limitations of LLMs}
%in Autonomous Mode}
\label{sec:limitations}

In this section, we will first review literature that calls into question claims about the planning and self-verification capabilities of LLMs. Subsequently, we will also provide some possible reasons for claims to the contrary made in the literature. 
\begin{table*}
    \centering
    \small

    \begin{tabular}{c c P{1.2cm} P{1.6cm} P{1.2cm} P{1.2cm} P{1.4cm}P{1.4cm}}
    \toprule
    \textbf{Domain} & \textbf{Method} & \multicolumn{6}{c}{
        \centering
        \textbf{Instances correct}}\\ \cmidrule{3-8}
        & & \textbf{GPT-4o} & \textbf{GPT-4-Turbo} & \textbf{Claude-3-Opus} & \textbf{LLaMA-3 70B} & \textbf{Gemini Pro} & \textbf{GPT-4} \\ \midrule[0.08em]
    \multicolumn{1}{P{2cm}}{\multirow{3}{=}{\textbf{Blocksworld (BW)}}} & One-shot & 170/600 (28.33\%) & 138/600 (23\%) & 289/600 (48.17\%) & 76/600 (12.6\%) & 68/600 (11.3\%) & 206/600 (34.3\%)  \\ \cmidrule{2-8}
    & Zero-shot & 213/600 (35.5\%) & 241/600 (40.1\%) & 356/600 (59.3\%) & 205/600 (34.16\%) & 3/600 (0.5\%) & 210/600 (34.6\%)\\ \cmidrule{2-8}
    \multicolumn{1}{P{2cm}}{\multirow{2}{=}{\textbf{Mystery BW (Deceptive)}}} & One-shot & 5/600 (0.83\%) & 5/600 (0.83\%) & 8/600 (1.3\%) & 15/600 (2.5\%) & 2/500 (0.4\%) &26/600 (4.3\%) \\ \cmidrule{2-8}
    & Zero-shot & 0/600 (0\%) & 1/600 (0.16\%) & 0/600 (0\%) & 0/600 (0\%) & (0/500) (0\%) & 1/600 (0.16\%)\\  
    \hline
    \bottomrule     
    \end{tabular}
    \caption{Results of state-of-the-art LLMs GPT-4o, GPT-4-Turbo, Claude-3-Opus, Gemini Pro and LLaMA-3 70B for Plan Generation with prompts in natural language.}
    \label{tab:planbench-sota}
\end{table*}
\subsection{LLMs cannot generate executable plans in autonomous mode}

%\rao{Good point; but will avoid this for now--as I am not too sure we are authorities on how humans plan.. Lucas: It may be good to make connections to how humans do planning (iteratively), and to problems that are intractable for combinatorial search.
% For example in my temporal planning draft I say:\\
% As planning domains become more expressive, even very advanced algorithms can struggle. In contrast, humans seem capable of solving problems that are intractable when specified formally.
% Our working theory of human planning accounts for this in two ways:
% \begin{enumerate}
%     \item The real world is highly sparse, in the sense that many states and actions are irrelevant to everyday life. Accordingly, human-like intuition plays a critical role in real-world planning. Heuristic candidate plan generation addresses this.
%     \item Humans plan iteratively by generating a candidate plan and refining it by responding to critiques. This avoids full-scale combinatorial search, but still solves real-world problems. LLM-Modulo with external verifiers addresses this.
% \end{enumerate}
% }

Despite initial claims about the planning capabilities of LLMs
\cite{bairi2023codeplan, yao2023react, shinn2023reflexion,
  huang2022inner, hao2023reasoning} several recent studies
confirm that LLMs are not actually able to generate executable plans
when they are used in autonomous modes
\cite{valmeekam2023on,liu2023llm+,silver2022pddl}. For example, in
\cite{valmeekam2023on,valmeekam2023planbench}, we evaluate LLMs' ability to generate correct plans on a suite of planning problem instances
% \footnote{Link to the github repo: \url{ https://github.com/karthikv792/gpt-plan-benchmark}}
%Thus, in this paper, we want to look at the ability of large language models to do reasoning about actions and change involving common-sense planning tasks. We develop a suite of benchmarks,
based on the kinds of domains employed in the International Planning Competition \cite{ipc}. 
% The tasks in the benchmark suite are aimed to test a variety of plan generation and validation capabilities. 
To eliminate the subjective aspect of analysis that forms the core part
of many earlier efforts to evaluate the reasoning capabilities of
LLMs, we leverage models and tools from the automated planning community to automate evaluation. 
%and tools to generate the queries and validate the system's answers. 
% While our primary interest is in plan generation, the test tasks themselves form a broad curriculum for evaluating LLM's capabilities of reasoning about actions and change.

We show that results in the autonomous mode are pretty bleak. On
average, only about 12\% of the plans that the best LLM (GPT-4)
generates are actually executable without errors and 
goal-reaching. We show that the choice of LLM doesn't have much
bearing on this. We tested the
family of GPT LLMs including GPT-4 \cite{openai2023gpt4}, GPT-3.5
\cite{openai_chatgpt}, InstructGPT-3 \cite{ouyang2022training} and GPT-3
\cite{brown2020language}.
%, as well as the latest ones including Claude Opus, Gemini, GPT4-Turbo and GPT4-o. 
We also show that fine-tuning does not seem to have a major effect on this dismal performance. 
We demonstrate that the performance deteriorates further if the names of the actions and objects in the domain are obfuscated--a change that doesn't in any way affect the performance of the standard AI planners. This further suggests that LLMs are more likely doing approximate retrieval of plans than actual planning. 
%To shed further light on the performance of GPT4, we present an evaluation of the plans it generates under a series of more relaxed (more forgiving) executability conditions.}
%
%Further, we provide a human baseline for the simplest domain in our set of domains, by presenting the planning instances to human subjects (through IRB-approved studies) and evaluating the quality and correctness of their plans.  These results are \textit{substantially better} than those of LLMs--confirming that LLMs can't plan even in a simple common sense domain in the autonomous mode. 
%

\bnote{We continue to reconfirm these limitations over each of the more
recently released LLMs, including Claude Opus, Gemini, GPT4-Turbo and
GPT4-o. Table~\ref{tab:planbench-sota} shows that all the
state of the art LLMs show dismal performance on PlanBench
\cite{valmeekam2023planbench}.}

\bnote{More recently, we have also investigated
so-called ``chain of thought'' prompting \cite{stechly2024chain}, as well as ReAct-style step-by-step prompting \cite{mudit-siddhant-react} and found that they too are largely
ineffective in improving the planning performance of 
LLMs.}

\subsection{LLMs cannot verify plans and thus cannot improve by self-critiquing }
\label{sec:cantverify}

There still exists considerable optimism that even if LLMs can't generate correct solutions in one go, their accuracy might improve in an iterative prompting regime, where LLMs will be able to ``self-critique" their candidate solutions and refine them to the point of correctness \cite{yao2023react,yao2023tree,shinn2023reflexion, weng2023large,huang2022inner}. 
%\rao{OK. Lucas: I'd say ``accuracy might improve" to be clearer in distinguishing hypothetical from empirical.}
This belief seems to rest largely on the assumption that verification of correctness should be easier than generation for many reasoning problems--a rather classical argument from computational complexity. There are grounds to be skeptical of this assumption as the complexity of the reasoning task should 
be irrelevant to LLM performance if what they are doing is approximate retrieval. 
In general, unless LLMs are trained not just on ``{\em correct data}," but also on ``{\em corrections data}," there is no {\em a priori} reason to believe that their critiques would even be approximately relevant, let alone actually correct. 
%\rao{Will stick to the hedged version--a la kaya. Lucas: ``Grounds to be skeptical" is hedging slightly. It is a stylistic choice, but another (in my opinion clearer) option is to say that self-critiquing simply doesn't work (outside of minor syntax issues), and to cite Kaya/Matthew/Karthik's paper.}
%\rao{Kaya: re Lucas minor quibble there is that we have empirical results that show this, but personally I'm not convinced that this is going to be true in every domain. Results like the reversal curse might imply that in some domains where there is just way more verification than generation data, the former is easier than the latter. A priori though, I expect that kind of asymmetry to be rare, so any given domain should be as easy to generate as to verify for an LLM.}

Two of our studies--one on plan verification \cite{valmeekam2023can} and the other on CSP verification \cite{stechly2023gpt} seem to throw cold water on this optimism. 
In \cite{stechly2023gpt}, we systematically investigate the effectiveness of iterative prompting in the context of {\em Graph Coloring}, a canonical NP-complete reasoning problem. 
%We chose graph coloring as it is representative of both of standard classes of reasoning problems studied in AI--propositional satisfiability and constraint satisfaction--and practical problems like scheduling and allocation. 
Our  methodology involves a principled empirical study of the
performance of GPT4 on two tasks: solving a large suite of random graph
coloring instances and, separately, verifying the correctness of the
candidate colorings--both in direct (i.e., return the first solution
generated by the LLM) and iterative modes. In iterative modes, we
experiment both with an LLM critiquing LLM-produced solutions and an
external, guaranteed correct reasoner verifying solutions. In both
cases, we analyze whether the content of criticisms actually affects
bottom-line performance. A more recent paper further analyzes these
results along with performance on the 24 puzzle--a task that has been used
by some researchers claiming LLMs have the ability to self verify
\cite{karthik-kaya-self-verification-combined}.

%\rao{Done Lin: seems that ``direct mode" is not defined or explained here}

Our results indicate that in direct mode, LLMs are, perhaps not surprisingly, pretty bad at solving graph coloring instances. More interestingly, they are no better at verifying solutions. In iterative modes, given the inability of LLMs to verify solutions, it should come as no surprise that our experiments also show that the strategy of LLMs self-critiquing their solutions does not improve over the baseline. We report that the performance is in fact {\em worse} because the system can't recognize a correct coloring and thus merrily passes over fortuitously correct colorings it has generated, ending up with a wrong one! Similar results have also been reported for planning problems in \cite{valmeekam2023on}. 

%\rao{self-improvement via fine-tuning itself on plans it generates and critiques also won't work}

One important corollary of the fact that LLMs cannot self-critique their plans is that they also can't self-improve by generating synthetic data, e.g. by generating plans themselves, critiquing the plans by themselves to improve them, and then using those to fine-tune themselves, as has been claimed in the literature \cite{huang-etal-2023-large, wang2022self}\footnote{Contrary to their claim of ``self-improvement", works like \cite{wang2022self} actually heavily depend on external knowledge (crafted seed examples) and critics (filtering step).}.%\cite{huang-etal-2023-large}.

\subsection{Analyzing Claims to the Contrary in the Literature}
\label{sec:contrary}
%\rao{need to add citations--react, reflect, verification claims}

Given that LLMs can neither guarantee correct generation nor correct verification of plans, as discussed in the previous sections, one obvious question is why the literature is replete with claims contrary to this
\cite{bairi2023codeplan, yao2023react, shinn2023reflexion,yao2023tree,weng2023large,huang2022inner}. 

\noindent{\bf Claims about Planning:} To analyze planning claims, we need to first understand that solving planning tasks requires (a) having the necessary planning domain knowledge–the actions and their preconditions, effects; the standard hierarchical recipes (e.g. task reduction schemas in HTN planning), past cases/plans, etc., and (b) being able to assemble this planning knowledge into an executable plan that takes care of any subgoal/resource interactions. The first part can be called knowledge acquisition and the second reasoning/planning. On closer examination, many papers claiming LLMs have planning abilities wind up confusing general planning knowledge extracted from the LLMs for executable plans. When all we are looking for are abstract plans, such as ``wedding plans," with no intention of actually executing them, it is easy to confuse them for complete executable plans.
%\rao{I think it is best to not get into exactly what is missing.. grounding can be done post-facto as in SayCAN Lucas: This is a great point, and I'd elaborate that \textit{execution requires grounding}. Since GPT has presumably never attended a wedding, it's likely it's never experienced the ways wedding plans can go wrong (or generally how they are executed). }
%
Indeed, our close examination of several works claiming planning
capabilities for LLMs \cite{llm-tutorial} suggests that they either work
in domains/tasks where subgoal interactions can be safely ignored
\cite{yao2023react, shinn2023reflexion}\footnote{Although domains like
AlfWorld \cite{ALFWorld20} do have sub-goal interactions for successful
task completion, \cite{yao2023react} and \cite{shinn2023reflexion}
largely ignore these interactions by either focusing on single subgoals
or relying on the ergodic nature of the domain
when prompting LLMs for generating plans \cite{mudit-siddhant-react}.}--either because they are
just working on a single subgoal, or because the world is forgiving and
ergodic; or by delegating the interaction resolution (reasoning)  to the
humans in the loop (who, through repeated prompting, have to ``correct"
the plan). Sometimes, in common sense domains, or with enough
fine-tuning, the ``assembling" part may also be obviated by having seen
a case that pretty much corresponds to the problem that needs to be
solved. Not surprisingly, our work  \cite{valmeekam2023on} shows that if the action interactions are removed by relaxing the world models, then the ability of LLMs to guess executable plans improves. Without these assumptions or mitigations, the plans that come out of LLMs may look reasonable to the lay user, and yet lead to execution time interactions and errors.\footnote{These issues are illustrated in part by a recent news story \cite{nyt-travel-books} about the proliferation of travel planning books, mostly auto-extracted from LLMs, and the ensuing disappointment of the unsuspecting end users who buy them mistaking them for usable plans!}

\begin{figure}
    \centering
    \includegraphics[width=\linewidth]{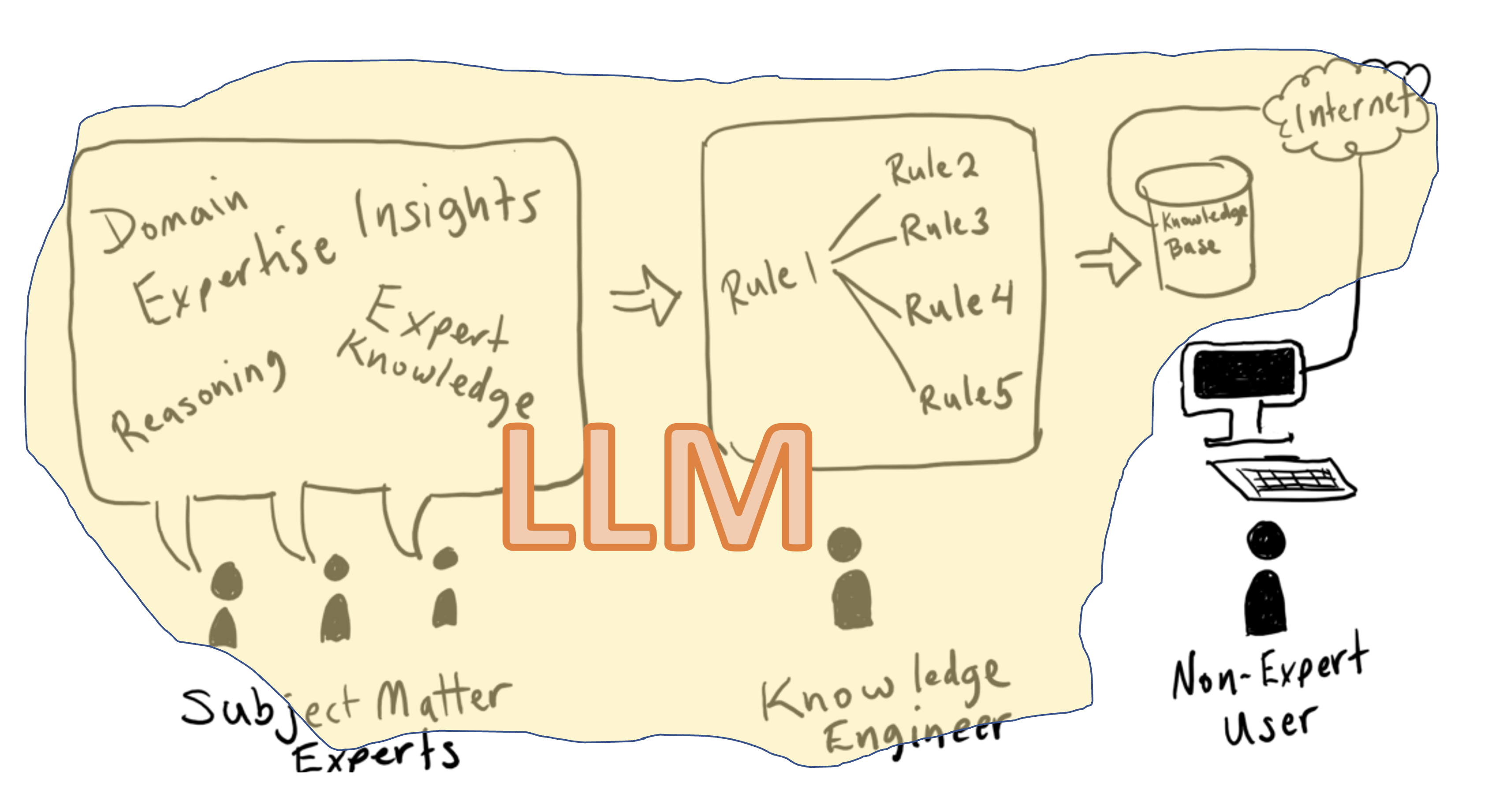}
    \caption{Viewing LLMs as an approximate knowledge source trained over civilizational knowledge}
    \vspace*{-0.2in}
    \label{fig:avenging-polanyi}
\end{figure}
%answer. 
%\rao{Lucas: "We believe.." is modest, consider making it stronger e.g. "LLM-Modulo is currently the most principled method for using LLMs in planning.". You could still say "We believe" beforehand to tone it down slightly, but I think "principled" gets the point across more clearly.}

%Traditional approaches to model-based reasoning/planning that focus on the incompleteness and incorrectness of the said models (such as model-lite planning \cite{zhuo2017model}, robust planning \cite{nguyen2017robust}) can have fresh relevance.

\begin{figure*}[h]
    \centering
    \includegraphics[width=\linewidth]{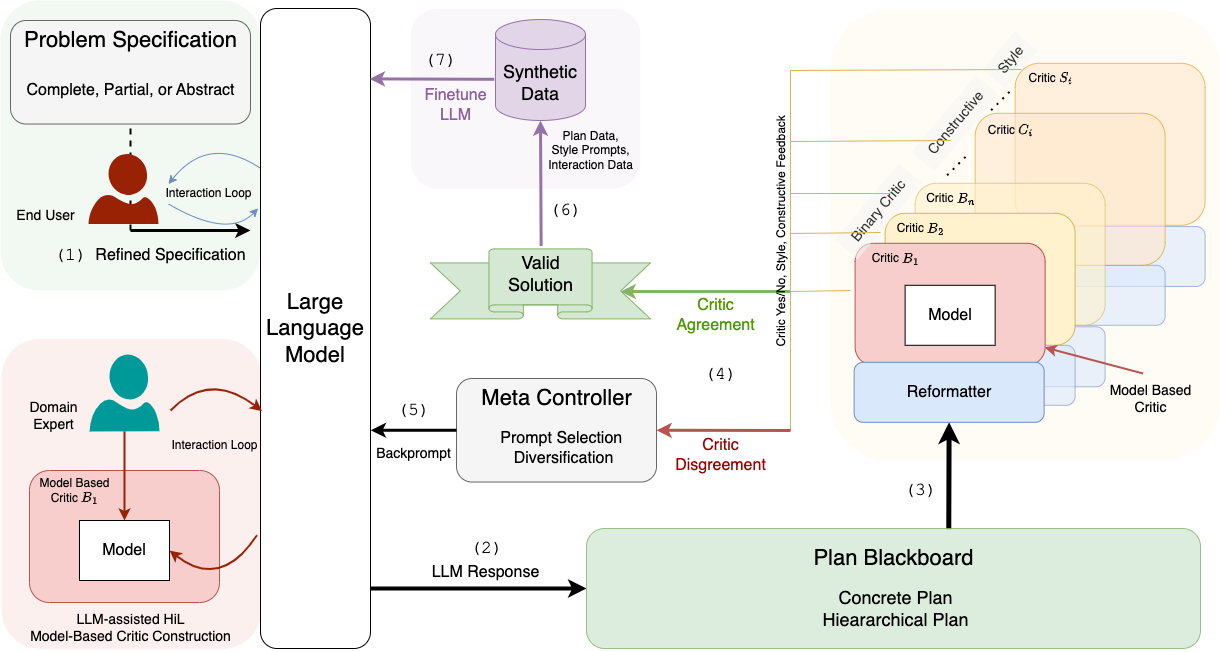}
    \caption{The proposed LLM-Modulo framework where LLMs act as idea generators and various external critics that specialize in different aspects, critique the candidate plan.}
    \label{fig:llm-modulo}
\end{figure*}

\noindent{\bf Claims about Self-Verification:} Coming to the claims about LLM's self-verification abilities, a closer look at the literature \cite{yao2023tree,huang2023large} shows that those claims are either (i) made in the context of tacit knowledge tasks for which there is little possibility of a verifier (e.g. essay writing)--making it hard to evaluate whether LLM's critiquing actually helped
%\footnote{Paradoxically, the fact that it is infeasible to write sound verifiers for tacit knowledge tasks also makes it possible for everyone to be a critic. Think of R2 saying the paper could be made "less dense" or the Peloton instructor critiquing Christopher Nolan film.}
or (ii) the external verification is carried out either by simulators \cite{wang2023voyager, yao2023react} or simple calls to the underlying operating system. %(as is the case, for example, for the 24 puzzle in \cite{yao2023tree}). 

%\donote{Kaya: you should check this part to make sure this gels with your read}
In a related vein, there is the recent Tree of Thoughts (ToT) paper \cite{yao2023tree}, which has been pitched as a way to convert LLMs into some type of systematic search with self-verification. 
%A closer look at the work however shows that ToT simply iteratively back-prompts the LLM until it comes up with a solution that is acceptable to an external verifier. 
%
Specifically, ToT employs a problem-specific prompt priming method. The ``tree" in ToT is essentially a way to generate diverse priming prompts (that the authors set up in a {\em problem specific way}). In other words, despite the use of terminology of problem-solving agents \cite{russell2010artificial}--search tree, expansion etc., there is really no deeper connection to search-based agents.%\footnote{Our preliminary experiments also show that at least in 24 puzzle, a simple iterative prompting, even without a systematic prompt diversification, is quite competitive with the ToT framework.}

% Lucas : Disregard the below
% \rao{Lucas: Based on the ToT discussions I've overheard, it may be ideal to explicitly say ToT isn't better than regeneration, assuming we don't want to hedge.}

%In other words, there is no obvious connection between

%Once you realize this, you will stop making questionable connections between ToT and search/reasoning!

%verification

The guarantees--if any--are coming in terms of soundness of the external verifier. The one clear reasoning problem used in the ToT paper is the 24 puzzle--for which the external verifier can be easily implemented in terms of arithmetic operations (thankfully not done by the numerically challenged LLM!). Here, our experiments show that the
%at least for problems with formal correctness criteria (e.g. 24 puzzle that the paper uses) the
LLM's own criticisms are often quite off the mark.\footnote{Note that we can do this check easily because of the formal specification of correctness. For the ``improving writing task" also used in ToT, there are no formal quality metrics and so it is hard to say anything concrete about the critiques of the LLM.}
Because the 24 puzzle's solutions can be verified by simple arithmetic operations, it is trivial to implement an external verifier for these problems.
%readers don't quite realize that the framework \textit{relies on an external verifier}.
In general though, the verifier may be more complex and can involve substantial work (you can substitute a simulator for the verifier--but someone has to write that simulator too!).
%\rao{Lucas: I added italics in the above to emphasize your point}

%
%\bnote{Commenting this out}
%In general planning problems, one way to provide an external verifier is to (a) write a domain model (e.g. in PDDL) and (b) feed it to an off-the-shelf model-based verifier like VAL (c.f. \cite{howey2004val}).  
%
%Finally, while there are claims that LLM critiques the intermediate solutions, we see that at least for problems with formal correctness criteria (e.g. 24 puzzle that the paper uses) the criticisms are often quite off the mark.

%In the case of ToT, the back-prompting to the LLM is guided more by the step-by-step hand-coded instructions from the human about diversifiying the prompts, rather than by the flaws found by the verifier. 

%\rao{should this para go elsewhere? Also add citations}

%\rao{we concede that there are different types of planning}

%\rao{verification claims are made for tasks without any possibility of formal verifiers.. LLMs for teaching! ToT criticism on the style}

%hoopla 

%Put the CACM blog stuff 

%Put ToT stuff here? 

%\subsection{Planning Background}

%\rao{take it from the NeuriPS paper}

%\section{LLM-Modulo Framework for Robust Planning: A Proposal}

\noindent{\bf LLMs as Approximate Knowledge Sources:}
The fact that LLMs are often good at extracting planning knowledge can indeed be gainfully leveraged. As shown in recent works \cite{guan2023leveraging}, LLMs can be a rich source of approximate models of world/domain dynamics and user preferences, as long as the humans (and any specialized critics) in the loop verify and refine those models, and give them over to model-based solvers. This way of using LLMs has the advantage that the humans need only be present when the dynamics/preference model is being teased out and refined, and the actual planning after that can be left to  
sounder planning frameworks with correctness guarantees, such as the LLM-Modulo framework we propose.

Such an overall approach has striking similarities to knowledge-based AI systems of yore, with LLMs effectively replacing the ``knowledge engineer" (see Figure~\ref{fig:avenging-polanyi}). Given the rather quixotic and dogmatic shift of AI away from approaches that accept domain knowledge from human experts that can be termed ``Polanyi's Revenge" (c.f. \cite{polanyi-revenge}),
%\rao{thx! Lin: ``I bemoaned in Polanyi's", will this violate double blind reviewing?}, 
this new trend of using LLMs as knowledge sources can be viewed as a form of avenging Polanyi's revenge!  Indeed, LLMs make it easy to get problem-specific knowledge as long as we are willing to relax the correctness requirements of that knowledge. In contrast to the old knowledge engineering approaches, LLMs offer this without making it look like we are inconveniencing any specific human (we are, instead,  just leveraging everything humans told each other on the Web!).  So the million dollar question for reasoning tasks is: ``{\em how would you do robust planning if you have some doddering know-it-all ready to give you any kind of knowledge?}" The LLM-Modulo Framework is a principled method for tackling this challenge. 

\section{LLM-Modulo Framework for Robust Planning}
\label{sec:llm_modulo_framework}

While Section~\ref{sec:limitations} questions the claims that LLMs are capable of planning/reasoning by themselves, it is certainly not meant to imply that LLMs don't have any constructive roles to play in solving planning/reasoning tasks. On the contrary, as discussed in the Introduction, their uncanny ability to generate ideas/potential candidate solutions–albeit with no guarantees about those guesses–can be valuable in the generate-test-critique setups in conjunction with either model-based verifiers or 
%planners, external solvers or 
expert humans in the loop. Accordingly, we propose a general ``LLM-Modulo'' framework\footnote{The name LLM-Modulo is inspired by the SAT-Modulo theories \cite{sat-modulo}.}. While we believe that versions of such an architecture can be of use in a wide variety of planning or reasoning tasks, for the sake of concreteness, we will focus on planning tasks, especially of the type studied in the automated planning community \cite{ghallab2004automated}.

Figure~\ref{fig:llm-modulo} gives a schematic of the {\bf LLM-Modulo Framework}, as we envision it. As can be seen readily, the underlying architecture is a Generate-Test-Critique loop, with the LLM generating candidate plans and a bank of critics critiquing the candidate. The loop starts with the LLM getting the problem specification and generating its first plan candidate.\footnote{Although we focus on planning from scratch, it is easy to accommodate replanning scenarios, where the loop starts with an externally supplied candidate plan.} 
%for planning tasks that facilitates these uses.  
Note that the plans an LLM helps generate in this architecture have {\em
  soundness guarantees}  because of the external sound critics. This
means that plans coming out of such an compound system  will constitute a
better corpus of synthetic data for any fine tuning phase carried out to
improve/customize the LLM's generation capability. The {\em
  completeness} of the system depends on the LLM's ability to generate
all potentially relevant candidates. 

%This in turn depends on how 

%While teacher-forced training seems to lead to incompleteness anyways \cite{vaishnav}

%Cite Vaishnav
%

\noindent{\bf Design Choices:} Before going into the details about the framework and its various modules, it is worth noting some design decisions underlying the proposed architecture. We start by noting that the LLM-Modulo architecture is a ``Generate-Test" one that involves LLMs interacting with the external \textbf{critics/verifiers} rather than a LLMs being just front-ends to external solvers. This is a deliberate decision--as this allows the LLM to guess/generate candidates to satisfy the critics, as against dealing with the expressiveness and search complexity issues of the solvers. \bnote{The critics/verifiers also are also more naturally {\em composable} than solvers/planners. As we shall see, we do allow for {\em constructive critics} which can be based on solvers, and provide suggestions on specific ways of extending/modifying the candidate plans.}

Secondly, the framework explicitly recognizes that the LLMs can generate approximate ideas not just about plan candidates, but domain models, problem reduction strategies, and refinements to the problem specification. The framework also recognizes that LLMs are good at format/syntax changes. Accordingly, the framework leverages all these abilities of LLMs, letting them play multiple roles in planning. 
Finally, the architecture carefully circumscribes the human's role--domain experts interact with the LLM to tease out the models used by (some of) the critics, while end users take part in refining any incomplete problem specification in concert with the LLM. A notable, and deliberate, absence is human's involvement in the inner loop of planning--e.g. with iterative prompting. 
In addition to posing an infeasible burden on the human's time for complex planning problems, such iterative prompting strategies are notorious for their Clever Hans effect \cite{cleverhans-wikipedia}.
%\rao{Lucas: This section is great}
%

%Don't use the planner but use the verifier
%    Don't make the human be involved in the search
%      iterative prompting
%
%   Recognize that LLMs can generate ideas not just about plans but
%     problem reduction strategies, domain models, problem specification
%     refinement etc
%Verification is indeed easier than generation for humans

%even if you want to fine tune on synthetic data.. 

%\subsection{Overview of the Architecture}

%The underlying architecture is a Generate-Test loop, with the LLM generating candidate plans and a bank of critics critiquing the candidate. 

%Three layers: (1) The plan guessing/back-prompting loop (2) LLMs helping with the models driving the critics (3) LLMs being finetuned with the plans generated this way

\subsection{Critics/Verifers}
%\rao{Siddhant NEW TEXT (rewriting the first paragraph):

In the LLM-Modulo framework, critics can evaluate LLM-generated
candidates for a planning/reasoning problem over both hard and soft (style)
constraints. Hard constraints refer to correctness verification which
can include causal correctness, timeline correctness, resource
constraint correctness as well as unit tests.
%\rao{Not clear who is doing the refinement.. will skip for now Mudit: If the LLM guess is a high level plan, appropriate critics which can check for constraint violations at an abstract level can be used to iteratively refine the blue print to a concrete plan.} 
For PDDL planning problems, the hard critic can be based on VAL
\cite{howey2004val}, that works off of a model (which itself can be
acquired with the help of the LLM \cite{guan2023leveraging}.
\bnote{It is worth noting that the critics don't always have to be declarative
model-based ones, and can be simulators. Just as LLMs can help humans in
coming up with models, they can also help in writing procedural
simulators, as seems to be done in systems like Voyager \cite{voyager}.} 
%NVIDIA system..

On the other hand, soft constraints can include more abstract notions of
good form such as style, explicability, preference conformance, etc.
As discussed in Section~\ref{sec:contrary}, while LLMs cannot take on
the role of hard critics with soundness guarantees,\footnote{If we don't insist on soundness guarantees, then it is, in principle, possible to train LLMs discriminatively to learn to verify plans; see \cite{arora2023learning}.} they can help simulate some aspects of the
role of soft (style) critics. So our framework does allow for style
critics be possibly based on LLMs. \bnote{For example, in
  \cite{verma2024theory} we discuss how LLMs can act as a human proxy to
  evaluate plans in terms of how they would be perceived by humans in
  the loop. Additionally, in \cite{guan2024task}, we show how Vision-Language
  Models (VLMs) can be leveraged to critique the style of robot
  behaviors in terms of their adherence to the soft common-sense
  preferences of the humans in the loop. }
%(e.g \cite{verma2024theory}). %\rao{Mudit: 
We reiterate that the soundness of the LLM-modulo framework is inherited
from the soundness of the correctness (hard) critics.

%Given the large corpora of data they are trained on, LLMs have shown remarkable performance in language understanding and inference tasks, thus making them viable for critiquing the conformance of a robot's plan to an expected behavior type (e.g. \cite{verma2024theory}).}

% \rao{Siddhant: Clarifying what we mean by `soft' constraints in the context of LLM-Modulo architecture can be helpful here, since currently, it does not seem very obvious for the kind of distinction we are trying to make. One way to address this could be to talk about what kind of information we are expecting from the soft critic and why this is more reasonable to expect from an LLM than correctness verification, hence the need for the hard critics a.k.a. ground-truth verification modules like VAL.}

%Critics can range from those that check for the hard constraints of correctness--causal correctness, timeline correctness, resource constraint correctness etc., to those that check for soft constraints such as style, explicability, preference conformance etc. As discussed in Section~\ref{sec:contrary}, while LLMs cannot take on the role of hard critics, they can help simulate some soft critics (e.g. \cite{verma2024theory}). For PDDL planning problems, the hard critic can be based on VAL \cite{howey2004val}, which works off of a model (which itself can be acquired with the help of the LLM \cite{guan2023leveraging}.

%\rao{Siddhant NEW TEXT: As shown in Figure \ref{fig:llm-modulo}, we also discuss the case where both hard and soft-constraint critics can be utilized in conjunction with each other.} 

The bank of critics--hard (model-based) as well as soft (possibly
LLM-based) evaluate the current plan candidate to evaluate its
fitness/acceptability. If at least all the hard critics sign off on the
current candidate, then that is considered a valid solution to be
returned to the end-user or the executor. When a critic finds the
current plan candidate to be unsatisfactory, it can provide varying
levels of feedback, ranging from ``{\em No, try again}" to ``{\em No,
  try again, here is one thing wrong with the current plan}" to ``{\em
  No, try again, here are all the things wrong with the current
  plan}."
\bnote{More importantly, the critics can be {\bf constructive}, and offer
alternatives plan/subplan suggestions. One way of obtaining such
constructive critics is to base them on partial planners--operating either on
the models themselves or their relaxations
\cite{pg-heuristics-aimag}.}
%These feedbacks are essentially 
%
These critiques are all pooled at the Meta (Backprompt)  Controller
(see Section~\ref{sec:meta-controller})

%Style critics based on observed human preferences vs. content/executability critics. 

%\subsubsection{LLMs as Reformulators}
\noindent{\bf LLMs as Reformulators:}
One interesting challenge is that many of the symbolic model-based
verifiers tend to be operating on specialized formal
representations. Given a central candidate plan (e.g. a mission plan),
these critics need translations of that candidate into their
representations. This is the role of the  reformulator module attached
to individual critics. These reformulator modules can be supported to a large extent by LLMs,
%, is to convert the central plan into specialized representations (or "views") used by the different critics.
given that one thing LLMs are very good at is format change across different syntactic representations \cite{olmo2021gpt3}.
%they can be used to allow for a symbolic \textit{lingua franca}. 
%
Indeed, as discussed in Section~\ref{sec:intro}, some approaches to
combine LLMs with external symbolic solvers just use LLMs as
reformulators for these solvers \cite{liu2023llm+,pan2023logic}.
\bnote{It is worth noting that the syntax conversion itself can be helped with
a nested LLM-Modulo framework--where the syntactic correctness of the
conversion is checked by syntax critics. We will have occasion to
illustrate this in the context of our preliminary work on  LLM-Modulo
frameworks for travel planning discussed in
Section~\ref{sec:case-studies}. }
%ChuChu Fan? 
Our discussion of LLM-Modulo framework should make it clear that syntax
reformulation alone is a severely limited role for LLMs!
%\rao{Lucas: I like this section, I also clarified to say "problem reformulation alone". You could also elaborate to re-mention tight integration of LLM Modulo and its various components}

%\rao{Lucas: Consider writing the above in terms of \textit{syntax}, e.g. ``LLMs seem empirically capable of following a formal grammar, and transforming knowledge into different syntactic forms."}

\subsection{Backprompt (Meta) Controller}
\label{sec:meta-controller}
%(better name?)}

The critiques from the various critics are pooled together by the Meta (Backprompt) Controller, which passes a processed version of them to the LLM as the next iterative prompt to elicit the next guess. 
%\rao{Mudit: 
This is especially required in the presence of a mix of soft and hard critics, where the Meta Controller can assume the responsibility of compiling the critiques into a consistent feedback to send to the LLM.

The processing in the controller can range from (i) simple round-robin selection of prompts to (ii) generating a summarized prompt (with LLM help) to (iii) employing a {\em prompt diversification strategy} to elicit the next candidate from a different part of the implicit search space. This last strategy 
%should be seen as an explicit effort to 
helps increase the completeness of the LLM candidate generation, and may involve domain/task-specific knowledge (see the discussion of Tree of Thoughts in Section~\ref{sec:contrary}). 

%(As discussed in \ref{sec:contrary}), this is effectively what Tree of Thoughts does \cite{yao2023tree}.) 

%Such summarization is a reasonable strategy as LLMs don't have the capability to treat back prompts  treated as hard constraints by LLMs anyway \cite{stechly2024chain}

%\rao{Since LLMs don't consider critiques as hard constraints, and written mostly to diversify

%\rao{Not sure right now how to include this. Anil: Prompt diversification strategies other than ToT - If necessary for citation purposes - include varying the temperature and scoring the prompts \& re-prompting LLM to generate better prompts - which is used in Google Deepmind's "LLM as Optimizers" paper }

%\rao{How about the problem decomposition as provided by humans as in ToT?}
%Diversify the prompt \cite{ToT} Mention that simple re-prompting is competitive with manually diversified prompting {\em a la} ToT. 

%\input{fine-tuning-and-synthetic-data-section}
\subsection{Specification Refinement \& Critic/Model Acquisition}
%%(Semi-automated)}

%\fromrao{the LLM knowledge avenging polanyi figure***}

%\donote{mention that specification itself can be llm-modulo..; also selecting crtics; agentic llms}

As mentioned earlier, we avoid having humans involved in iteratively prompting LLMs--as this can be an infeasibly time-consuming activity for them. Instead, we let automated verifiers, either model-based or LLM-supported, to manage the plan critiquing process. 
The framework does depend on humans for ``once per domain" and ``once
per problem" interactions. In the former category, human domain experts
can play a role in acquiring the domain model with the help of the
LLM. Examples of such interaction include teasing out PDDL planning
models from the LLMs with the help of human expert curation (top left in
Figure~\ref{fig:llm-modulo}). An example of this is our work in 
\cite{guan2023leveraging}. The idea here is that the traditional domain model acquisition task (e.g. \cite{simpson2001gipo}) is significantly made easier by having the LLMs help with ideas regarding various pieces of the domain model (e.g., actions, their preconditions and effects) and letting humans sign off/critique the resulting model. Once the model is acquired this way, it can be used by correctness verifiers such as VAL \cite{howey2004val,guan2023leveraging}.
Often the planning problems in real world situations are specified
incompletely, leaving it to the human commonsense to refine the
specification. This brings up a second role for humans--this time end
users (bottom left in Figure~\ref{fig:llm-modulo}--in collaboratively
refining the specification with the help of LLMs (similar to the way
done in  \cite{xie2023translating,liu2023llm+}).

%\donote{Issue of specification vs. Syntax; also enumerating potential critics}

\subsection{Summary of LLM Roles in LLM-Modulo}
%Framework}
%%
It is worth summarizing the  multiple roles the LLM plays in the LLM-Modulo architecture. The most prominent, of course, is its role in ``guessing'' the candidate plans (step 2 in Figure~\ref{fig:llm-modulo}) in response to problem specification and iterative back prompting from the bank of critics (Step 5). 
Second, the LLM plays a role in converting the guessed plan candidate into specialized representations used by the various critics (e.g., the time-line view, the causal link view etc.). This role leverages the fact that LLMs are very good at format conversion (c.f. \cite{olmo2021gpt3}). 
Third, the LLM plays a role in helping the end user flesh out the incomplete problem specification to begin with (Step 1 in Figure~\ref{fig:llm-modulo}). 
Finally, the LLM plays a role in helping the domain expert tease out and refine the domain models used by the various model-based critics \cite{guan2023leveraging,kwon2022reward}, or help ``implement" procedural critics (such as those checking syntactic constraints). \bnote{As a broad approximate source of knowledge,  the LLM can also help enumerate the list of potential critics needed to validate the candidate plans (once again with a human in the loop).}

\begin{figure}
    \centering
    \includegraphics[width=\linewidth]{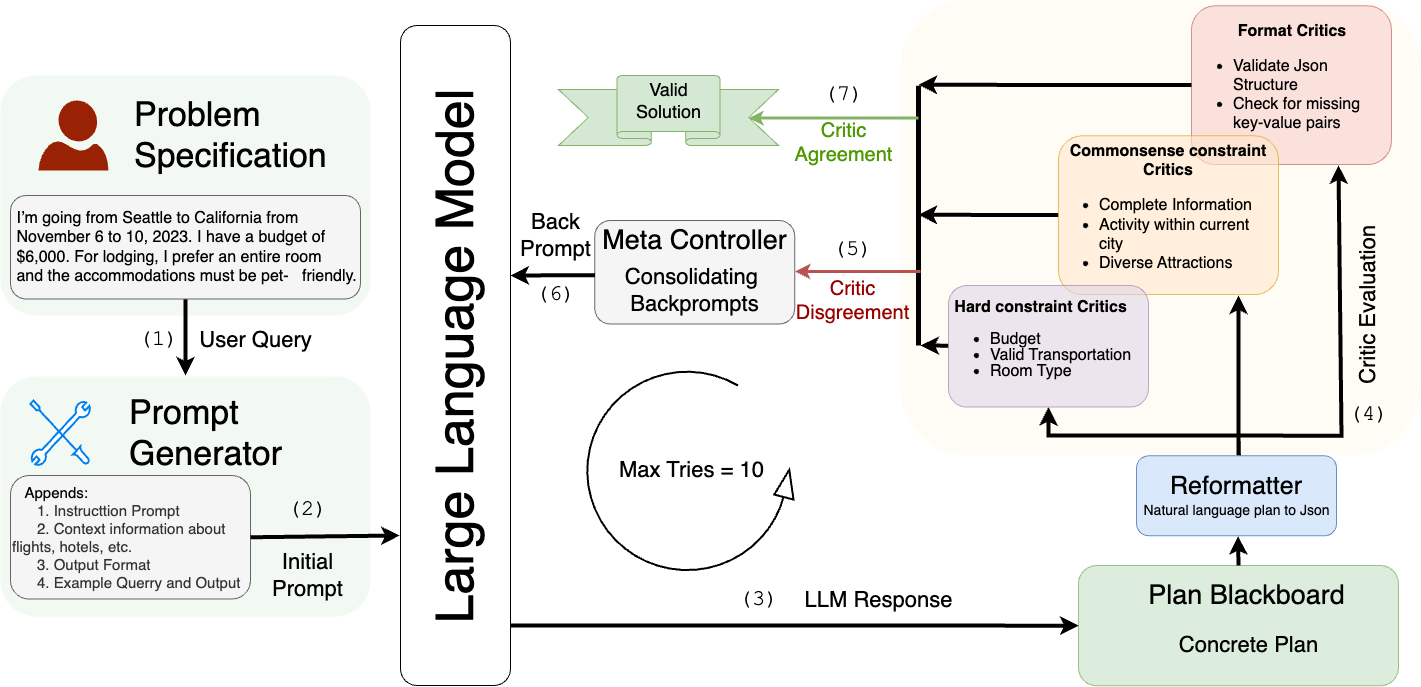}
    \caption{LLM Modulo Framework adapted for Travel Planning}
    \label{fig:llm_modulo_TP}
    \vspace*{-0.1in}
\end{figure}

\begin{figure}
    \centering
    \includegraphics[width=0.9\linewidth]{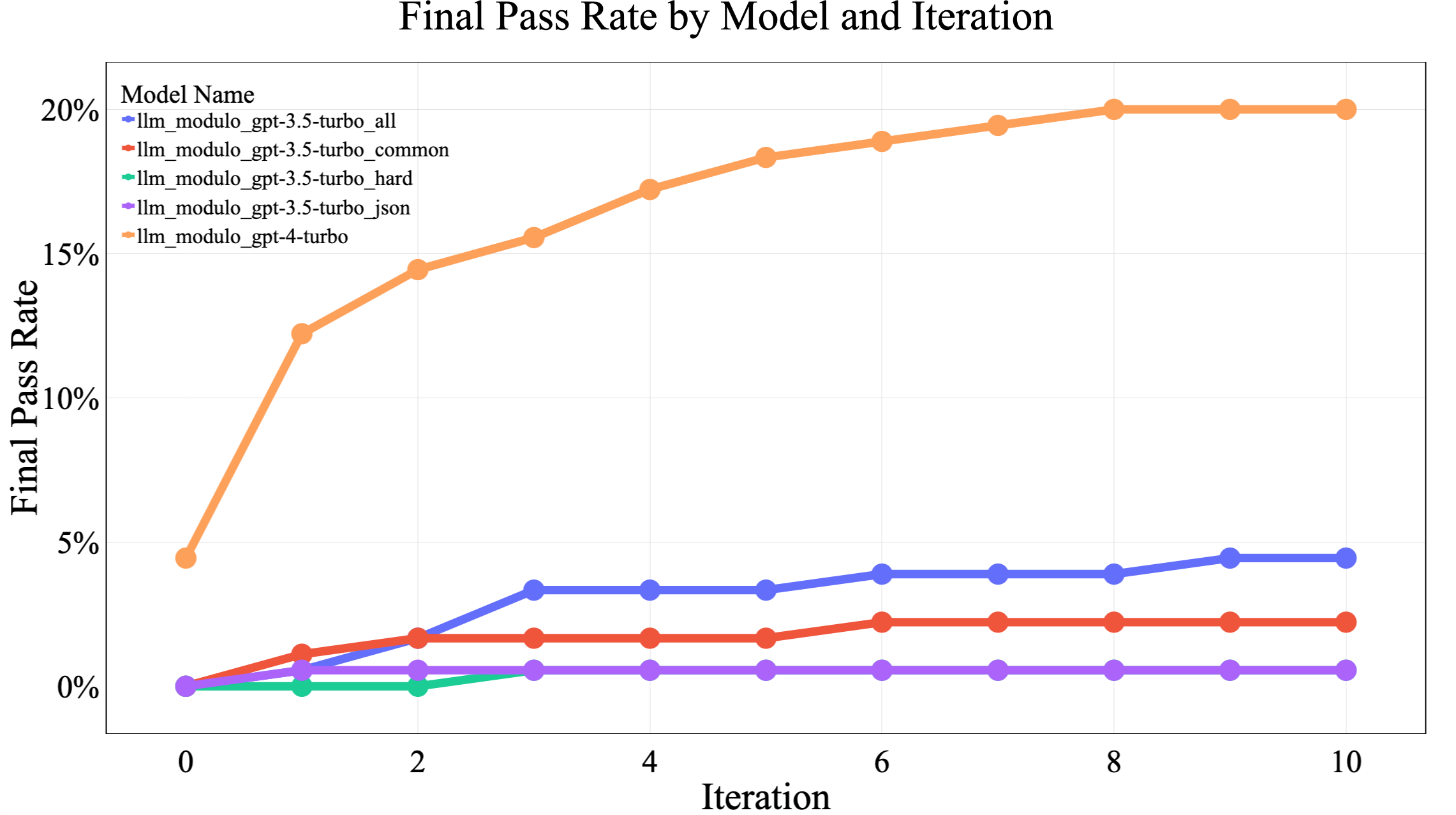}
    \caption{Final Pass rates of models across LLM Modulo Iterations}
    \vspace*{-0.2in}
    \label{fig:final_pass_rate_iter}
\end{figure}

\subsection{Can LLM-Modulo Frameworks Pay Their Way?}
\label{sec:shoe-horn}
%One question regarding this LLM-Modulo architecture from a planning perspective is whether 

Let's address the elephant in the room: {\em Given that formal model-based planning systems already exist \cite{ghallab2004automated}, is LLM-Modulo framework for planning more than a gratuitous attempt to shoe-horn (the currently popular) LLMs to solve planning problems?}
%, when there are more formal combinatorial planning systems available in multiple communities \cite{ghallab2004automated}. 
\bnote{ Indeed, when the underlying problem is actually solvable by such
  combinatorial solvers, it can be orders of magnitue more resource
  efficient to use them.\footnote{\bnote{Not surprisingly, automated
  programming, the one community
  that certainly doesn't have the luxury of a ready-made ``solver,''
  have stuck to LLM-Modulo style approaches.}}
%In short, to what extent is LLM-Modulo architecture for planning is like using LLMs as calculators. 
}
Compared to a planner that is guaranteed to be correct in a narrow set of domains, LLMs may likely be good at generating plausible (but not guaranteed to be correct) plan heuristics/suggestions in many more scenarios. 
Thus, unlike the traditional planning architectures studied in AI \cite{ghallab2004automated}, which put \textit{a priori} constraints on the expressiveness of the problems that can be posed to the planner (to wit, the different expressiveness levels of the PDDL specification \cite{mcdermott1998pddl}), the LLM-Modulo architecture puts no such restrictions. 
In this sense, it is more representative of real-world planning problems
such as those in NASA mission planning, where the different
critics--human and automated--are at best able to give ``no objection"
certificates for the candidate plans under consideration,  clearing it from their perspective.
(Indeed, both deep space network planning and mars rover task planning are done via a \textit{collective human blackboard}. \cite{johnston2014automated,bresina2004activity}.) 
%without any global guarantees of success. 
Note that this is starkly different from just sending an unvetted plan out to execution (as would be the case if we have LLMs operate in autonomous mode to guess plans). Generalizing planning and reasoning frameworks this way is consistent with the Doyle \& Patil's call to the Knowledge Representation community of yore \cite{twotheses-kr}, as well as our own call for model-lite planning \cite{rao-model-lite}.
\section{Two Case Studies of LLM-Modulo}
\label{sec:case-studies}
We have applied the LLM-Modulo framework to classical planning domains (as reported in \cite{valmeekam2023on}) and to a recent travel planning benchmark (as reported in \cite{llm-modulo-travel-planning}). In the former case, the results (presented in Section 5.2 and Table 4 of \cite{valmeekam2023on}) show that with back prompting from VAL \cite{howey2004val} acting as the external verifier and critic, LLM performance in Blocks World improves to 82\% within 15 back prompting rounds, while in Logistics, it improves to 70\%. LLM-Modulo doesn't help as much in an obfuscated version of blocks world called Mystery BW, reaching about 10\% accuracy. This should be expected because the LLMs have difficulty generating plausible candidate plans for this domain (note that even here, if a plan is returned, it must have passed muster with VAL, and is thus guaranteed correct by its model).

For the travel planning case study, we used a benchmark proposed in \cite{xie2024travelplanner}, which involves a rich mix of travel constraints presented in flexible natural language form. Our preliminary results on adapting LLM-Modulo framework to this benchmark are reported in \cite{llm-modulo-travel-planning}. 
% Authors of the benchmark themselves provide variations of agentic LLMs such as GPT models paired with prompt engineering techniques like Chain of Thought and ReAct, and report that the best LLM strategies exhibit a startlingly low 0.7\% performance rate!
The benchmark's authors test LLMs across a variety of prompt engineering techniques including Chain of Thought and ReAct, reporting that--on GPT-3.5-Turbo--the current best strategies only manage a startlingly low 0.7\% performance rate!
%Unlike other reasoning benchmarks, TravelPlanner highlights the faliure of existing agentification of LLMs with extremely low performance rates.
We adapted the LLM-Modulo framework to this benchmark by operationalizing their hard constraints (such as the budget constraint set by the user) or common-sense constraints (such as suggesting diverse attractions to visit) as critics as shown in Figure~\ref{fig:llm_modulo_TP}.  Our preliminary results show (see Figure~\ref{fig:final_pass_rate_iter}; additional results in \cite{llm-modulo-travel-planning}) that LLM-Modulo based agentification with automated critics in the loop significantly improves the performance (6x of baselines) even with a limit of 10 back prompting cycles, and weaker models such as GPT-3.5-turbo. Furthermore, we also find that LLMs can successfully implement functions corresponding to hard critics and several common-sense critics. Finally, LLMs reliably play the role of reformatter as well, converting free form travel plans into structured plans parseable by the critics for backprompts or plan evaluation. One interesting observation about this domain is that we were able to use the LLM itself to enumerate the type of critics needed to validate the plan (with light human supervision). 

%\donote{Results on blocks world and travel planning domain}

\section{Related Work}

%We will note that 
While the LLM-Modulo framework is being proposed in general form here for the first time, there are certainly works in leveraging LLMs in planning and reasoning tasks that are in line with the spirit of the LLM-Modulo framework. 
%For exmaple, both \cite{valmeekam2023on} and \cite{stechly2023gpt}  describe and evaluate a backprompting interaction between an LLM and an external verifier. 
Work on FunSearch \cite{funsearch} depends on a generate-test loop between a specially fine-tuned LLM that guesses solutions, and an external symbolic evaluator that critiques them. The authors note how the external verifier is critical for avoiding falling prey to hallucinations (i.e., approximate solution candidates that have flaws). AlphaGeometry \cite{alphageometry} too depends on the Generate-Test-Critique interaction between a fine-tuned LLM and a symbolic evaluator. Both these systems fine-tune pre-trained LLMs with task specific synthetic data--the correctness of which is vetted with external simulators.
%(as we discuss in Section~\ref{sec:syntheticdata}). 
%Even the earliest work using LLMs 

%Alpha Geometry and FunSearch--have a lot in common with LLM-Modulo. 

While we focused on PDDL planning tasks for the sake of concreteness, we believe that the essence of LLM-Modulo framework is equally applicable to other scenarios involving planning and reasoning--such as \textit{Reinforcement Learning with Simulators}.  Such RL systems rely on rewards as feedback to train a policy. 
Simulators takes on the roles of 
%Analogous to the 
plan evaluation and critiques performed by the respective critics in the LLM-Modulo framework (e.g. \cite{alvaro-saynav}).
%, environment simulators provide the required feedback. 
The fact that simulators play the role of verifiers is often
%largely overlooked 
not explicitly recognized in cases where LLMs are used as an actor to generate an admissible plan by interacting with a simulator, for example in the case of AlfWorld \cite{yao2023react, shinn2023reflexion} and Minecraft \cite{wang2023voyager}.
%
%As mentioned in Section \ref{sec:llm_modulo_framework}, 
Similar to extracting a domain model such as in the case of \cite{guan2023leveraging}, LLMs can also be used for designing a reward model or shaping the reward
\cite{siddhant-llm-rl,kwon2022reward, hao2023reasoning,ma2023eureka}.
%for the plan generation - feedback cycle is yet another potential use case that has been recently looked at, 
%for reward shaping 

%}, and for text-based } and robot manipulation \cite{ma2023eureka} domains.

Interestingly, the fact that LLM's can help come up with approximate quasi-symbolic transition models, reward models and models of high level actions has made a bigger splash in RL. This is because for far too long, researchers there have tried to spurn any high level models (lest that would involve depending on humans; \cite{polanyi-revenge}) and focused on learning to act from sensory information, under the name of ``deep reinforcement learning." Given the horrendous sample complexity of the DRL methods even in reaching a single subgoal, and the well known fact that even approximate symbolic models can help drastically improve the performance (c.f. \cite{lin-icml22}), coupled with the fact that LLM's are only too glad to dream up approximate models and goal recipes, there has been a performance revolution of sorts there \cite{yao2023react,liang2023code,wang2023voyager}. If we look beyond the improvements in these lower level goal seeking behaviors--especially in the presence of ergodic simulators, the RL approaches dependent on LLMs will encounter the same issues regarding subgoal interactions that our discussion of PDDL planning problems brought into focus. The LLM-Modulo inspired frameworks will thus, we believe, be equally relevant there. Indeed, SayCan \cite{ahn2022saycan} the earliest use of LLMs in generating policies in an RL-with-Simulator scenario, explicitly filters the action choices suggested by the LLM with the help of simulator. 

%Lin work approximate models 

%RLHF is the only survivor. 
Although we focused on text based LLMs (such as GPT4), recently there have also been impressive development in multi-modal LLMs (e.g. GPT4V). 
%\cite{GPT4v,Bard}. 
While multi-modality is a great addition that increases the coverage of their System 1 imagination (Figure~\ref{fig:sys12}), 
%\rao{Mudit: and potentially allows for a larger context to work with}, 
it is not clear that this gives them System 2 competence.\footnote{If you know how to complete sentences, and now learned to complete dance moves, does your ability to reason/plan magically improve?} 
As we
discussed earlier, we can leverage VLMs for style criticism of the robot
behavior \cite{guan2024task}.

\bnote{Finally, our position (with published supporting evidence) that LLMs are incapable
of supporting planning in autonomous modes must seem quite at odds with
the current head-long rush into agentic LLMs. We believe that the latter is
largely a result of confusing ``acting'' with ``planning.''
Given their ability to translate across formalisms, it 
is of course possible for LLMs to invoke external services--something
frameworks like AutoGPT and LangChain support. But the mere ability to
invoke an action doesn't, in any way, guarantee that the course of
actions thus invoked will achieve a desired state of affairs. The only
way to guarantee the latter is to to support robust planning capabilities--something
our LLM-Modulo frameworks strive to do.}

%Langchain autogpt references

%\section{Summary \& Conclusion}

\section{Conclusion}

This 
position 
paper is a modest attempt to combat both over-optimism and over-pessimism about the role of LLMs in planning and reasoning tasks. 
Our position is that \textit{LLMs cannot plan themselves but can play a variety of constructive roles in solving planning tasks--especially as approximate knowledge sources and candidate plan generators in the so-called LLM-Modulo Frameworks in conjunction with external sound model-based verifiers.} In support of this position, we summarized the literature questioning the claims about the planning and self-verification capabilities of LLMs by themselves. We also discussed how conflating approximate knowledge acquisition and generating executable plans of action is behind many of the claims about planning and verification abilities of LLMs. We then shared LLM-Modulo framework, our vision for a productive way to integrate the impressive idea generation/approximate knowledge provision capabilities of LLMs with external verifiers with correctness guarantees, for robust and expressive planning. We discussed how planning in LLM-Modulo framework avoids inheriting the expressiveness and search-complexity limitations of traditional symbolic planners, while retaining their soundness guarantees. We illustrated and discussed the many roles LLMs can play in the LLM-Modulo framework. Finally, we also discussed two case studies of adapting the LLM-Modulo frameworks.  
%As we discussed, LLM-Modulo frameworks are consistent with some of the most high-profile success stories of ``neuro-symbolic" architectures, including AlphaGeometry and FunSearch. 
\newpage

\section*{Impact Statement}

% \section*{Impact Statement}
This position paper takes a stance on a robust and well-founded way of leveraging Large Language Models in planning and reasoning tasks. It (i) points out the inabilities of current pre-trained LLMs to tackle planning problems, (ii) suggests some reasons as to why there are wide-spread misunderstandings about LLM planning abilities and (iii) proposes LLM-Modulo frameworks as a way to leverage LLMs for robust planning. The main consequences of realizing this position/vision is expected to be (i) sounding caution about misapplication of LLMs in autonomous modes for planning (ii) providing a way to leverage LLMs to do robust planning. 
Given the current interest in agentic LLMs, these insights 
can have significant positive impact in mission critical situations. We do not see any obvious negative societal consequences of leveraging LLMs this way (unless of course the plans are aimed at achieving malicious goals). 

%work whose goal is to advance the field of Machine Learning. There are many potential societal consequences of our work, none which we feel must be specifically highlighted here.

\section*{Acknowledgments}
The ideas discussed in this paper have evolved over a series of talks, tutorials and twitter threads. The discussions, feedback and encouragement from colleagues, including Sarath Sreedharan, Tom Dietterich, %Kathleen Fisher, 
Yann LeCun, Daniel Borrajo, and Dan Weld  
is gratefully acknowledged. 
The adaptation of LLM-Modulo Framework for Travel Planning, discussed in Section~\ref{sec:case-studies} was lead by Atharva Gundawar.
Kambhampati acknowledges generous support from ONR via grants N00014-18-1-2442, N14-18-1-2840 and N00014-23-1-2409, as well as gifts from J.P. Morgan, Qualcomm and Amazon.
\bibliography{llmplan}
\bibliographystyle{icml2024}
%\bibliographystyle{apa}
%\newpage
%\input{0-todo-ideas}
\end{document}

%% file: commands.tex
\newcommand{\knote}{\color{red}}

%% file: main.bbl
\begin{thebibliography}{67}
\providecommand{\natexlab}[1]{#1}
\providecommand{\url}[1]{\texttt{#1}}
\expandafter\ifx\csname urlstyle\endcsname\relax
  \providecommand{\doi}[1]{doi: #1}\else
  \providecommand{\doi}{doi: \begingroup \urlstyle{rm}\Url}\fi

\bibitem[cle()]{cleverhans-wikipedia}
Clever {H}ans.
\newblock https://en.wikipedia.org/wiki/Clever\_Hans.

\bibitem[Ahn et~al.(2022)Ahn, Brohan, Brown, Chebotar, Cortes, David, Finn, Fu, Gopalakrishnan, Hausman, et~al.]{ahn2022saycan}
Ahn, M., Brohan, A., Brown, N., Chebotar, Y., Cortes, O., David, B., Finn, C., Fu, C., Gopalakrishnan, K., Hausman, K., et~al.
\newblock Do as i can, not as i say: Grounding language in robotic affordances.
\newblock \emph{arXiv preprint arXiv:2204.01691}, 2022.

\bibitem[Arora \& Kambhampati(2023)Arora and Kambhampati]{arora2023learning}
Arora, D. and Kambhampati, S.
\newblock Learning and leveraging verifiers to improve planning capabilities of pre-trained language models.
\newblock \emph{ICML Workshop on Knowledge and Logical Reasoning in the Era of Data-driven Learning (arXiv preprint arXiv:2305.17077)}, 2023.

\bibitem[Bairi et~al.(2023)Bairi, Sonwane, Kanade, Iyer, Parthasarathy, Rajamani, Ashok, Shet, et~al.]{bairi2023codeplan}
Bairi, R., Sonwane, A., Kanade, A., Iyer, A., Parthasarathy, S., Rajamani, S., Ashok, B., Shet, S., et~al.
\newblock Codeplan: Repository-level coding using llms and planning.
\newblock \emph{arXiv preprint arXiv:2309.12499}, 2023.

\bibitem[Bhambri et~al.(2024)Bhambri, Bhattacharjee, Liu, and Kambhampati]{siddhant-llm-rl}
Bhambri, S., Bhattacharjee, A., Liu, H., and Kambhampati, S.
\newblock Efficient reinforcement learning via large language model-based search, 2024.

\bibitem[Bresina et~al.(2004)Bresina, J{\'o}nsson, Morris, and Rajan]{bresina2004activity}
Bresina, J.~L., J{\'o}nsson, A.~K., Morris, P.~H., and Rajan, K.
\newblock Activity planning for the mars exploration rovers.
\newblock In \emph{ICAPS-2005 Conference}, 2004.

\bibitem[Brown et~al.(2020)Brown, Mann, Ryder, Subbiah, Kaplan, Dhariwal, Neelakantan, Shyam, Sastry, Askell, et~al.]{brown2020language}
Brown, T., Mann, B., Ryder, N., Subbiah, M., Kaplan, J.~D., Dhariwal, P., Neelakantan, A., Shyam, P., Sastry, G., Askell, A., et~al.
\newblock Language models are few-shot learners.
\newblock \emph{Advances in neural information processing systems}, 33:\penalty0 1877--1901, 2020.

\bibitem[Bryce \& Kambhampati(2007)Bryce and Kambhampati]{pg-heuristics-aimag}
Bryce, D. and Kambhampati, S.
\newblock A tutorial on planning graph based reachability heuristics.
\newblock \emph{{AI} Mag.}, 28\penalty0 (1):\penalty0 47--83, 2007.

\bibitem[Bubeck et~al.(2023)Bubeck, Chandrasekaran, Eldan, Gehrke, Horvitz, Kamar, Lee, Lee, Li, Lundberg, et~al.]{bubeck2023sparks}
Bubeck, S., Chandrasekaran, V., Eldan, R., Gehrke, J., Horvitz, E., Kamar, E., Lee, P., Lee, Y.~T., Li, Y., Lundberg, S., et~al.
\newblock Sparks of artificial general intelligence: Early experiments with gpt-4.
\newblock \emph{arXiv preprint arXiv:2303.12712}, 2023.

\bibitem[Doyle \& Patil(1991)Doyle and Patil]{twotheses-kr}
Doyle, J. and Patil, R.~S.
\newblock Two theses of knowledge representation: Language restrictions, taxonomic classification, and the utility of representation services.
\newblock \emph{Artificial intelligence}, 48\penalty0 (3):\penalty0 261--297, 1991.

\bibitem[Dziri et~al.(2023)Dziri, Lu, Sclar, Li, Jiang, Lin, Welleck, West, Bhagavatula, Bras, Hwang, Sanyal, Ren, Ettinger, Harchaoui, and Choi]{dziri2023faith}
Dziri, N., Lu, X., Sclar, M., Li, X.~L., Jiang, L., Lin, B.~Y., Welleck, S., West, P., Bhagavatula, C., Bras, R.~L., Hwang, J.~D., Sanyal, S., Ren, X., Ettinger, A., Harchaoui, Z., and Choi, Y.
\newblock Faith and fate: Limits of transformers on compositionality.
\newblock In \emph{Thirty-seventh Conference on Neural Information Processing Systems}, 2023.
\newblock URL \url{https://openreview.net/forum?id=Fkckkr3ya8}.

\bibitem[Gendron et~al.(2023)Gendron, Bao, Witbrock, and Dobbie]{gendron2023large}
Gendron, G., Bao, Q., Witbrock, M., and Dobbie, G.
\newblock Large language models are not abstract reasoners.
\newblock \emph{arXiv preprint arXiv:2305.19555}, 2023.

\bibitem[Ghallab et~al.(2004)Ghallab, Nau, and Traverso]{ghallab2004automated}
Ghallab, M., Nau, D., and Traverso, P.
\newblock \emph{Automated Planning: theory and practice}.
\newblock Elsevier, 2004.

\bibitem[Guan et~al.(2022)Guan, Sreedharan, and Kambhampati]{lin-icml22}
Guan, L., Sreedharan, S., and Kambhampati, S.
\newblock Leveraging approximate symbolic models for reinforcement learning via skill diversity.
\newblock In Chaudhuri, K., Jegelka, S., Song, L., Szepesvari, C., Niu, G., and Sabato, S. (eds.), \emph{Proceedings of the 39th International Conference on Machine Learning}, volume 162 of \emph{Proceedings of Machine Learning Research}, pp.\  7949--7967. PMLR, 17--23 Jul 2022.
\newblock URL \url{https://proceedings.mlr.press/v162/guan22c.html}.

\bibitem[Guan et~al.(2023)Guan, Valmeekam, Sreedharan, and Kambhampati]{guan2023leveraging}
Guan, L., Valmeekam, K., Sreedharan, S., and Kambhampati, S.
\newblock Leveraging pre-trained large language models to construct and utilize world models for model-based task planning.
\newblock In \emph{Thirty-seventh Conference on Neural Information Processing Systems}, 2023.
\newblock URL \url{https://openreview.net/forum?id=zDbsSscmuj}.

\bibitem[Guan et~al.(2024)Guan, Zhou, Liu, Zha, Amor, and Kambhampati]{guan2024task}
Guan, L., Zhou, Y., Liu, D., Zha, Y., Amor, H.~B., and Kambhampati, S.
\newblock "task success" is not enough: Investigating the use of video-language models as behavior critics for catching undesirable agent behaviors, 2024.

\bibitem[Gundawar et~al.(2024)Gundawar, Verma, Guan, Valmeekam, Bhambri, and Kambhampati]{llm-modulo-travel-planning}
Gundawar, A., Verma, M., Guan, L., Valmeekam, K., Bhambri, S., and Kambhampati, S.
\newblock Robust planning with llm-modulo framework: Case study in travel planning.
\newblock \emph{arXiv preprint arxiv:2405.20625}, 2024.

\bibitem[Hao et~al.(2023)Hao, Gu, Ma, Hong, Wang, Wang, and Hu]{hao2023reasoning}
Hao, S., Gu, Y., Ma, H., Hong, J.~J., Wang, Z., Wang, D.~Z., and Hu, Z.
\newblock Reasoning with language model is planning with world model.
\newblock \emph{arXiv preprint arXiv:2305.14992}, 2023.

\bibitem[Howey et~al.(2004)Howey, Long, and Fox]{howey2004val}
Howey, R., Long, D., and Fox, M.
\newblock {VAL: Automatic plan validation, continuous effects and mixed initiative planning using PDDL}.
\newblock In \emph{16th IEEE International Conference on Tools with Artificial Intelligence}, pp.\  294--301. IEEE, 2004.

\bibitem[Huang et~al.(2023{\natexlab{a}})Huang, Chen, Mishra, Zheng, Yu, Song, and Zhou]{huang2023large}
Huang, J., Chen, X., Mishra, S., Zheng, H.~S., Yu, A.~W., Song, X., and Zhou, D.
\newblock Large language models cannot self-correct reasoning yet.
\newblock \emph{arXiv preprint arXiv:2310.01798}, 2023{\natexlab{a}}.

\bibitem[Huang et~al.(2023{\natexlab{b}})Huang, Gu, Hou, Wu, Wang, Yu, and Han]{huang-etal-2023-large}
Huang, J., Gu, S., Hou, L., Wu, Y., Wang, X., Yu, H., and Han, J.
\newblock Large language models can self-improve.
\newblock In Bouamor, H., Pino, J., and Bali, K. (eds.), \emph{Proceedings of the 2023 Conference on Empirical Methods in Natural Language Processing}, pp.\  1051--1068, Singapore, December 2023{\natexlab{b}}. Association for Computational Linguistics.
\newblock \doi{10.18653/v1/2023.emnlp-main.67}.
\newblock URL \url{https://aclanthology.org/2023.emnlp-main.67}.

\bibitem[Huang et~al.(2022)Huang, Xia, Xiao, Chan, Liang, Florence, Zeng, Tompson, Mordatch, Chebotar, et~al.]{huang2022inner}
Huang, W., Xia, F., Xiao, T., Chan, H., Liang, J., Florence, P., Zeng, A., Tompson, J., Mordatch, I., Chebotar, Y., et~al.
\newblock Inner monologue: Embodied reasoning through planning with language models.
\newblock \emph{arXiv preprint arXiv:2207.05608}, 2022.

\bibitem[IPC(1998)]{ipc}
IPC.
\newblock International planning competition, 1998.
\newblock URL \url{https://www.icaps-conference.org/competitions/}.

\bibitem[Johnston et~al.(2014)Johnston, Tran, Arroyo, Sorensen, Tay, Carruth, Coffman, and Wallace]{johnston2014automated}
Johnston, M.~D., Tran, D., Arroyo, B., Sorensen, S., Tay, P., Carruth, B., Coffman, A., and Wallace, M.
\newblock Automated scheduling for nasa's deep space network.
\newblock \emph{AI Magazine}, 35\penalty0 (4):\penalty0 7--25, 2014.

\bibitem[Kahneman(2011)]{thinking-fast-slow}
Kahneman, D.
\newblock \emph{Thinking, fast and slow}.
\newblock macmillan, 2011.

\bibitem[Kambhampati(2007)]{rao-model-lite}
Kambhampati, S.
\newblock Model-lite planning for the web age masses: The challenges of planning with incomplete and evolving domain models.
\newblock In \emph{Proceedings of the National Conference on Artificial Intelligence}, volume~22, pp.\  1601. Menlo Park, CA; Cambridge, MA; London; AAAI Press; MIT Press; 1999, 2007.

\bibitem[Kambhampati(2021)]{polanyi-revenge}
Kambhampati, S.
\newblock Polanyi's revenge and {AI}'s new romance with tacit knowledge.
\newblock \emph{Communications of the ACM}, 64\penalty0 (2):\penalty0 31--32, 2021.

\bibitem[Kambhampati(2024)]{rao-cacm}
Kambhampati, S.
\newblock Can {LLMs} reason and plan?
\newblock \emph{Annals of the New York Academy of Sciences}, 2024.

\bibitem[Kambhampati et~al.(2023)Kambhampati, Valmeekam, Marquez, and Guan]{llm-tutorial}
Kambhampati, S., Valmeekam, K., Marquez, M., and Guan, L.
\newblock On the role of large language models in planning, July 2023.
\newblock URL \url{https://yochan-lab.github.io/tutorial/ICAPS-2023/}.
\newblock Tutorial presented at the International Conference on Automated Planning and Scheduling (ICAPS), Prague.

\bibitem[Kugel \& Hiltner(2023)Kugel and Hiltner]{nyt-travel-books}
Kugel, S. and Hiltner, S.
\newblock A new frontier for travel scammers: {A.I.-Generated Guidebooks}.
\newblock \emph{New York Times}, August 2023.
\newblock URL \url{https://www.nytimes.com/2023/08/05/travel/amazon-guidebooks-artificial-intelligence.html}.

\bibitem[Kwon et~al.(2022)Kwon, Xie, Bullard, and Sadigh]{kwon2022reward}
Kwon, M., Xie, S.~M., Bullard, K., and Sadigh, D.
\newblock Reward design with language models.
\newblock In \emph{The Eleventh International Conference on Learning Representations}, 2022.

\bibitem[Liang et~al.(2023)Liang, Huang, Xia, Xu, Hausman, Ichter, Florence, and Zeng]{liang2023code}
Liang, J., Huang, W., Xia, F., Xu, P., Hausman, K., Ichter, B., Florence, P., and Zeng, A.
\newblock Code as policies: Language model programs for embodied control, 2023.

\bibitem[Liu et~al.(2023)Liu, Jiang, Zhang, Liu, Zhang, Biswas, and Stone]{liu2023llm+}
Liu, B., Jiang, Y., Zhang, X., Liu, Q., Zhang, S., Biswas, J., and Stone, P.
\newblock Llm+ p: Empowering large language models with optimal planning proficiency.
\newblock \emph{arXiv preprint arXiv:2304.11477}, 2023.

\bibitem[Ma et~al.(2023)Ma, Liang, Wang, Huang, Bastani, Jayaraman, Zhu, Fan, and Anandkumar]{ma2023eureka}
Ma, Y.~J., Liang, W., Wang, G., Huang, D.-A., Bastani, O., Jayaraman, D., Zhu, Y., Fan, L., and Anandkumar, A.
\newblock Eureka: Human-level reward design via coding large language models.
\newblock \emph{arXiv preprint arXiv:2310.12931}, 2023.

\bibitem[McCoy et~al.(2023)McCoy, Yao, Friedman, Hardy, and Griffiths]{mccoy2023embers}
McCoy, R.~T., Yao, S., Friedman, D., Hardy, M., and Griffiths, T.~L.
\newblock Embers of autoregression: Understanding large language models through the problem they are trained to solve.
\newblock \emph{arXiv preprint arXiv:2309.13638}, 2023.

\bibitem[McDermott et~al.(1998)McDermott, Ghallab, Howe, Knoblock, Ram, Veloso, Weld, and Wilkins]{mcdermott1998pddl}
McDermott, D., Ghallab, M., Howe, A.~E., Knoblock, C.~A., Ram, A., Veloso, M.~M., Weld, D.~S., and Wilkins, D.~E.
\newblock Pddl-the planning domain definition language.
\newblock 1998.

\bibitem[Nieuwenhuis \& Oliveras(2006)Nieuwenhuis and Oliveras]{sat-modulo}
Nieuwenhuis, R. and Oliveras, A.
\newblock On sat modulo theories and optimization problems.
\newblock In \emph{Theory and Applications of Satisfiability Testing-SAT 2006: 9th International Conference, Seattle, WA, USA, August 12-15, 2006. Proceedings 9}, pp.\  156--169. Springer, 2006.

\bibitem[Olmo et~al.(2021)Olmo, Sreedharan, and Kambhampati]{olmo2021gpt3}
Olmo, A., Sreedharan, S., and Kambhampati, S.
\newblock Gpt3-to-plan: Extracting plans from text using gpt-3.
\newblock \emph{FinPlan 2021}, pp.\ ~24, 2021.

\bibitem[OpenAI(2022)]{openai_chatgpt}
OpenAI.
\newblock Introducing chatgpt by openai, 2022.
\newblock URL \url{https://openai.com/blog/chatgpt}.

\bibitem[OpenAI(2023)]{openai2023gpt4}
OpenAI.
\newblock Gpt-4 technical report, 2023.

\bibitem[Ouyang et~al.(2022)Ouyang, Wu, Jiang, Almeida, Wainwright, Mishkin, Zhang, Agarwal, Slama, Ray, et~al.]{ouyang2022training}
Ouyang, L., Wu, J., Jiang, X., Almeida, D., Wainwright, C., Mishkin, P., Zhang, C., Agarwal, S., Slama, K., Ray, A., et~al.
\newblock Training language models to follow instructions with human feedback.
\newblock \emph{Advances in Neural Information Processing Systems}, 35:\penalty0 27730--27744, 2022.

\bibitem[Pan et~al.(2023)Pan, Albalak, Wang, and Wang]{pan2023logic}
Pan, L., Albalak, A., Wang, X., and Wang, W.~Y.
\newblock Logic-lm: Empowering large language models with symbolic solvers for faithful logical reasoning.
\newblock \emph{arXiv preprint arXiv:2305.12295}, 2023.

\bibitem[Rajvanshi et~al.(2023)Rajvanshi, Sikka, Lin, Lee, Chiu, and Velasquez]{alvaro-saynav}
Rajvanshi, A., Sikka, K., Lin, X., Lee, B., Chiu, H.-P., and Velasquez, A.
\newblock Saynav: Grounding large language models for dynamic planning to navigation in new environments.
\newblock \emph{arXiv preprint arXiv:2309.04077}, 2023.

\bibitem[Romera-Paredes et~al.(2023)Romera-Paredes, Barekatain, Novikov, Balog, Kumar, Dupont, Ruiz, Ellenberg, Wang, Fawzi, et~al.]{funsearch}
Romera-Paredes, B., Barekatain, M., Novikov, A., Balog, M., Kumar, M.~P., Dupont, E., Ruiz, F.~J., Ellenberg, J.~S., Wang, P., Fawzi, O., et~al.
\newblock Mathematical discoveries from program search with large language models.
\newblock \emph{Nature}, pp.\  1--3, 2023.

\bibitem[Russell \& Norvig(2010)Russell and Norvig]{russell2010artificial}
Russell, S.~J. and Norvig, P.
\newblock \emph{Artificial intelligence a modern approach}.
\newblock London, 2010.

\bibitem[Shinn et~al.(2023)Shinn, Cassano, Gopinath, Narasimhan, and Yao]{shinn2023reflexion}
Shinn, N., Cassano, F., Gopinath, A., Narasimhan, K.~R., and Yao, S.
\newblock Reflexion: Language agents with verbal reinforcement learning.
\newblock In \emph{Thirty-seventh Conference on Neural Information Processing Systems}, 2023.

\bibitem[Shridhar et~al.(2021)Shridhar, Yuan, C\^ot\'e, Bisk, Trischler, and Hausknecht]{ALFWorld20}
Shridhar, M., Yuan, X., C\^ot\'e, M.-A., Bisk, Y., Trischler, A., and Hausknecht, M.
\newblock {ALFWorld: Aligning Text and Embodied Environments for Interactive Learning}.
\newblock In \emph{Proceedings of the International Conference on Learning Representations (ICLR)}, 2021.
\newblock URL \url{https://arxiv.org/abs/2010.03768}.

\bibitem[Silver et~al.(2022)Silver, Hariprasad, Shuttleworth, Kumar, Lozano-P{\'e}rez, and Kaelbling]{silver2022pddl}
Silver, T., Hariprasad, V., Shuttleworth, R.~S., Kumar, N., Lozano-P{\'e}rez, T., and Kaelbling, L.~P.
\newblock {PDDL} planning with pretrained large language models.
\newblock In \emph{NeurIPS 2022 Foundation Models for Decision Making Workshop}, 2022.
\newblock URL \url{https://openreview.net/forum?id=1QMMUB4zfl}.

\bibitem[Simpson et~al.(2001)Simpson, McCluskey, and Zhao]{simpson2001gipo}
Simpson, R., McCluskey, T.~L., and Zhao, W.
\newblock Gipo: an integrated graphical tool to support knowledge engineering in ai planning.
\newblock In \emph{ECP-01}, pp.\  445. Citeseer, 2001.

\bibitem[Stechly et~al.(2023)Stechly, Marquez, and Kambhampati]{stechly2023gpt}
Stechly, K., Marquez, M., and Kambhampati, S.
\newblock {GPT-4 Doesn’t Know It’s Wrong: An Analysis of Iterative Prompting for Reasoning Problems}.
\newblock In \emph{NeurIPS 2023 Foundation Models for Decision Making Workshop}, 2023.

\bibitem[Stechly et~al.(2024{\natexlab{a}})Stechly, Valmeekam, and Kambhampati]{karthik-kaya-self-verification-combined}
Stechly, K., Valmeekam, K., and Kambhampati, S.
\newblock On the self-verification limitations of large language models on reasoning and planning tasks.
\newblock \emph{arXiv preprint arxiv:2402.08115}, 2024{\natexlab{a}}.

\bibitem[Stechly et~al.(2024{\natexlab{b}})Stechly, Valmeekam, and Kambhampati]{stechly2024chain}
Stechly, K., Valmeekam, K., and Kambhampati, S.
\newblock Chain of thoughtlessness: An analysis of cot in planning.
\newblock \emph{arXiv preprint arxiv:2405.04776}, 2024{\natexlab{b}}.

\bibitem[Trinh et~al.(2024)Trinh, Wu, Le, He, and Luong]{alphageometry}
Trinh, T.~H., Wu, Y., Le, Q.~V., He, H., and Luong, T.
\newblock Solving olympiad geometry without human demonstrations.
\newblock \emph{Nature}, 625\penalty0 (7995):\penalty0 476--482, 2024.

\bibitem[Ullman(2023)]{ullman2023large}
Ullman, T.
\newblock Large language models fail on trivial alterations to theory-of-mind tasks.
\newblock \emph{arXiv preprint arXiv:2302.08399}, 2023.

\bibitem[Valmeekam et~al.(2023{\natexlab{a}})Valmeekam, Marquez, and Kambhampati]{valmeekam2023can}
Valmeekam, K., Marquez, M., and Kambhampati, S.
\newblock Can large language models really improve by self-critiquing their own plans?
\newblock In \emph{NeurIPS 2023 Foundation Models for Decision Making Workshop}, 2023{\natexlab{a}}.

\bibitem[Valmeekam et~al.(2023{\natexlab{b}})Valmeekam, Marquez, Olmo, Sreedharan, and Kambhampati]{valmeekam2023planbench}
Valmeekam, K., Marquez, M., Olmo, A., Sreedharan, S., and Kambhampati, S.
\newblock Planbench: An extensible benchmark for evaluating large language models on planning and reasoning about change.
\newblock In \emph{Thirty-seventh Conference on Neural Information Processing Systems Datasets and Benchmarks Track}, 2023{\natexlab{b}}.
\newblock URL \url{https://openreview.net/forum?id=YXogl4uQUO}.

\bibitem[Valmeekam et~al.(2023{\natexlab{c}})Valmeekam, Marquez, Sreedharan, and Kambhampati]{valmeekam2023on}
Valmeekam, K., Marquez, M., Sreedharan, S., and Kambhampati, S.
\newblock On the planning abilities of large language models - a critical investigation.
\newblock In \emph{Thirty-seventh Conference on Neural Information Processing Systems (Spotlight)}, 2023{\natexlab{c}}.
\newblock URL \url{https://openreview.net/forum?id=X6dEqXIsEW}.

\bibitem[Verma et~al.(2024{\natexlab{a}})Verma, Bhambri, and Kambhampati]{mudit-siddhant-react}
Verma, M., Bhambri, S., and Kambhampati, S.
\newblock On the brittle foundations of react prompting for agentic large language models.
\newblock \emph{arXiv preprint arXiv:2405.13966}, 2024{\natexlab{a}}.

\bibitem[Verma et~al.(2024{\natexlab{b}})Verma, Bhambri, and Kambhampati]{verma2024theory}
Verma, M., Bhambri, S., and Kambhampati, S.
\newblock Theory of mind abilities of large language models in human-robot interaction: An illusion?
\newblock \emph{arXiv preprint arXiv:2401.05302}, 2024{\natexlab{b}}.

\bibitem[Wang et~al.(2023{\natexlab{a}})Wang, Xie, Jiang, Mandlekar, Xiao, Zhu, Fan, and Anandkumar]{voyager}
Wang, G., Xie, Y., Jiang, Y., Mandlekar, A., Xiao, C., Zhu, Y., Fan, L., and Anandkumar, A.
\newblock Voyager: An open-ended embodied agent with large language models, 2023{\natexlab{a}}.

\bibitem[Wang et~al.(2023{\natexlab{b}})Wang, Xie, Jiang, Mandlekar, Xiao, Zhu, Fan, and Anandkumar]{wang2023voyager}
Wang, G., Xie, Y., Jiang, Y., Mandlekar, A., Xiao, C., Zhu, Y., Fan, L., and Anandkumar, A.
\newblock Voyager: An open-ended embodied agent with large language models.
\newblock \emph{arXiv preprint arXiv:2305.16291}, 2023{\natexlab{b}}.

\bibitem[Wang et~al.(2022)Wang, Kordi, Mishra, Liu, Smith, Khashabi, and Hajishirzi]{wang2022self}
Wang, Y., Kordi, Y., Mishra, S., Liu, A., Smith, N.~A., Khashabi, D., and Hajishirzi, H.
\newblock Self-instruct: Aligning language model with self generated instructions.
\newblock \emph{arXiv preprint arXiv:2212.10560}, 2022.

\bibitem[Weng et~al.(2023)Weng, Zhu, Xia, Li, He, Liu, Sun, Liu, and Zhao]{weng2023large}
Weng, Y., Zhu, M., Xia, F., Li, B., He, S., Liu, S., Sun, B., Liu, K., and Zhao, J.
\newblock Large language models are better reasoners with self-verification.
\newblock In \emph{Findings of the Association for Computational Linguistics: EMNLP 2023}, pp.\  2550--2575, 2023.

\bibitem[Xie et~al.(2024)Xie, Zhang, Chen, Zhu, Lou, Tian, Xiao, and Su]{xie2024travelplanner}
Xie, J., Zhang, K., Chen, J., Zhu, T., Lou, R., Tian, Y., Xiao, Y., and Su, Y.
\newblock Travelplanner: A benchmark for real-world planning with language agents.
\newblock \emph{arXiv preprint arxiv:2402.01622}, 2024.

\bibitem[Xie et~al.(2023)Xie, Yu, Zhu, Bai, Gong, and Soh]{xie2023translating}
Xie, Y., Yu, C., Zhu, T., Bai, J., Gong, Z., and Soh, H.
\newblock Translating natural language to planning goals with large-language models.
\newblock \emph{arXiv preprint arXiv:2302.05128}, 2023.

\bibitem[Yao et~al.(2023{\natexlab{a}})Yao, Yu, Zhao, Shafran, Griffiths, Cao, and Narasimhan]{yao2023tree}
Yao, S., Yu, D., Zhao, J., Shafran, I., Griffiths, T.~L., Cao, Y., and Narasimhan, K.~R.
\newblock Tree of thoughts: Deliberate problem solving with large language models.
\newblock In \emph{Thirty-seventh Conference on Neural Information Processing Systems}, 2023{\natexlab{a}}.
\newblock URL \url{https://openreview.net/forum?id=5Xc1ecxO1h}.

\bibitem[Yao et~al.(2023{\natexlab{b}})Yao, Zhao, Yu, Du, Shafran, Narasimhan, and Cao]{yao2023react}
Yao, S., Zhao, J., Yu, D., Du, N., Shafran, I., Narasimhan, K.~R., and Cao, Y.
\newblock React: Synergizing reasoning and acting in language models.
\newblock In \emph{The Eleventh International Conference on Learning Representations}, 2023{\natexlab{b}}.
\newblock URL \url{https://openreview.net/forum?id=WE_vluYUL-X}.

\end{thebibliography}
